\documentclass{article}
\usepackage{spconf,amsmath,graphicx,multirow,threeparttable,booktabs,amssymb}
\usepackage{scalerel,stackengine}
\usepackage{makecell}
\usepackage{booktabs}
\stackMath
\newcommand\reallywidehat[1]{%
	\savestack{\tmpbox}{\stretchto{%
			\scaleto{%
				\scalerel*[\widthof{\ensuremath{#1}}]{\kern-.6pt\bigwedge\kern-.6pt}%
				{\rule[-\textheight/2]{1ex}{\textheight}}
			}{\textheight}%
		}{0.5ex}}%
	\stackon[1pt]{#1}{\tmpbox}%
}

\title{An empirical study of domain-agnostic semi-supervised learning via energy-based models: joint-training and pre-training}
\name{Yunfu Song$^\dagger$, Huahuan Zheng$^\dagger$, Zhijian Ou\thanks{This work is supported by NSFC 61976122. $\dagger$ Equal contributions.}}
\address{Speech Processing and Machine Intelligence (SPMI) Lab, Tsinghua University, Beijing, China.\\
ozj@tsinghua.edu.cn
}
\begin{document}
\ninept
\maketitle
\begin{abstract}
A class of recent semi-supervised learning (SSL) methods heavily rely on domain-specific data augmentations.
In contrast, generative SSL methods involve unsupervised learning based on generative models by either joint-training or pre-training, and are more appealing from the perspective of being domain-agnostic, since they do not inherently require data augmentations.
Joint-training estimates the joint distribution of observations and labels, while pre-training is taken over observations only.
Recently, energy-based models (EBMs) have achieved promising results for generative modeling.
Joint-training via EBMs for SSL has been explored with encouraging results across different data modalities. In this paper, we make two contributions. First, we explore pre-training via EBMs for SSL and compare it to joint-training.
Second, a suite of experiments are conducted over domains of image classification and natural language labeling to give a realistic whole picture of the performances of EBM based SSL methods.
It is found that joint-training EBMs outperform pre-training EBMs marginally but nearly consistently.
\end{abstract}

\section{Introduction}
\label{sec:intro}

Deep neural networks (DNNs) have achieved great success in various domains mainly through supervised learning, which requires large collections of labeled data.
However, in many domains, collecting labeled data is difficult and expensive, but there are often easily-available unlabeled data.
This has motivated the community to develop semi-supervised learning (SSL), which aims to leveraging both labeled and unlabeled data for training.
There have emerged a plethora of SSL methods that are designed for deep neural networks \cite{miyato2018virtual,laine2016temporal,tarvainen2017mean,sohn2020fixmatch,chen2020simple,cvt}, spanning over domains of image classification, natural language labeling and so on.

The key to designing SSL methods is how to effectively exploit the information contained in the unlabeled data \cite{zhu2006semi}, which can provide regularizations for finding good classifiers. A Bayesian view of regularizations is priors, which reflect our priori knowledge regarding the model \cite{miyato2018virtual}.
Recent SSL methods with DNNs can be distinguished by the priors they adopt, and, roughly speaking, can be divided into two classes\footnote{We mainly discuss SSL methods for using DNNs.
	General discussion of SSL can be referred to \cite{zhu2006semi}.} - based on generative models or discriminative models. In this paper, we refer to these two classes as generative SSL and discriminative SSL respectively.
A popular priori used by discriminative SSL is that the outputs from the discriminative classifier are smooth with respect to local and random perturbations of the inputs. Examples include virtual adversarial training (VAT) \cite{miyato2018virtual} and a number of recently developed consistency-regularization based methods \cite{laine2016temporal,tarvainen2017mean,sohn2020fixmatch} and contrastive learning based methods \cite{chen2020simple}.
However, those SSL methods heavily rely on domain-specific data augmentations \cite{cubuk2020randaugment}, which are tuned intensively for images leading to impressive performance in some image domains but are less successful for other domains where these augmentations are less effective (e.g., medical images and text). For instance, random input perturbations are more difficult to apply to discrete data like text \cite{cvt}.

Generative SSL methods involve unsupervised learning over unlabeled data based on generative models.
As argued in \cite{bengio2013representation}, for observation $x$ and label $y$, learning $p(x)$ can be thought of providing a kind of generic priori to learning $p(y|x)$. Representations that are useful for $p(x)$ tend to be useful when learning $p(y|x)$, allowing sharing of statistical strength between the unsupervised and supervised learning.
In this sense, generative SSL methods is more appealing from the perspective of being domain-agnostic, since they do not inherently require data augmentations and generally can be applied to a wider range of domains.

Remarkably, there exist two different manners for the generative SSL approach - joint-training and pre-training.
In the first manner, which often referred to as joint-training, a joint model of $p(x,y)$ is defined.
When we have label $y$, we maximize $p(y|x)$ (the supervised objective), and when the label is unobserved, we marginalize it out and maximize $p(x)$ (the unsupervised objective). Semi-supervised learning over a mix of labeled and unlabeled data is formulated as maximizing the (weighted) sum of $\log p(y|x)$ and $\log p(x)$.
In the pre-training manner of SSL, we perform unsupervised representation learning on unlabeled data followed by supervised training (called fine-tuning) on labeled data.
This manner of pre-training followed by fine-tuning has received increasing application in natural language processing. 

This paper focuses on pushing forward domain-agnostic semi-supervised learning, particularly via energy-based generative models.
Recently, energy-based models (EBMs) \cite{pami,nrf,IGG-EBM,xie2016theory} have achieved promising results for generative modeling.
Joint-training via EBMs for SSL has been explored with very encouraging results \cite{nrf,zhaojoint,jrf}, which show state-of-the-art SSL performance across different data modalities (images, natural languages, an protein structure prediction and year prediction from the UCI dataset repository) and in different data settings (fix-dimensional and sequence data).
However, pre-training via EBMs for SSL has not been studied, and it is interesting to compare joint-training and pre-training when both are based on EBMs.
In this paper, we make two contributions. First, we explore pre-training via EBMs for SSL and compare it to joint-training.
Second, as suggested in \cite{oliver2018realistic}, a suite of experiments are conducted\footnote{Code will be released for reproduction upon acceptance of this paper.}, in which we vary both the amount of labeled and unlabeled data to give a realistic whole picture of the performances of EBM based SSL methods.

\section{Related work}
\label{sec:related-work}

\textbf{~~~~Discriminative and generative SSL.}
Semi-supervised learning is a heavily studied problem, ranging from classic self-training \cite{scudder1965probability} (also known as pseudo-labeling \cite{lee2013pseudo}), graph based methods \cite{zhu2003semi}, to various recent SSL methods for using DNNs.
In general, recent DNN based SSL methods can be distinguished by the prior they adopt for representation learning from unlabeled data.
Discriminative SSL works by discriminating between different augmentations from a given unlabeled sample, such as in recent FixMatch \cite{sohn2020fixmatch}, SimCLR \cite{chen2020simple} methods. They rely on a rich set of domain-specific data augmentations, e.g., RandAugment \cite{cubuk2020randaugment}. Although there are some efforts to use data-independent model noises, e.g., by dropout \cite{srivastava2014dropout}, domain-specific data augmentations is indispensable.

Recent progress in learning with deep generative models stimulates the generative SSL research, which usually involves blending unsupervised learning and supervised learning. These methods make fewer domain-specific assumptions and tend to be domain-agnostic.
The performance comparisons between generative and discriminative SSL methods are mixed.
It is found that consistency based discriminative SSL methods often outperform generative SSL methods in image domain.
However, in text domain, the generative SSL methods such as those based on pre-training word vectors are more successful and widely used.

\textbf{EBM based generative SSL.}
Recently, it is shown in \cite{nrf} that joint-training via EBMs produce state-of-the-art SSL results on images (MNIST, SVHN and CIFAR-10), compared to previous  generative SSL methods based on VAEs and GANs.
It is also shown in \cite{zhaojoint} that joint-training via EBMs outperforms VAT on tabular data from the UCI dataset repository other than images.
Further, joint-training via EBMs has been extended to modeling sequences and consistently outperform conditional random fields (CRFs) (the supervised baseline) and self-training (the classic semi-supervised baseline) on natural language labeling tasks such as POS tagging, chunking and NER.

On the other hand, pre-training has received attention in the early stage of training DNNs and recently become widely used in natural language processing tasks.
Pre-training via EBMs is conceptually feasible, but remain unexplored.
Despite the encouraging results obtained by joint-training of EBMs, it is not clear whether pre-training via EBMs is also competitive for domain-agnostic SSL.


\section{Semi-supervised Learning via EBMs}
\subsection{Background}
\label{sec:background}

An energy-based model (EBM) \cite{Lecun}, also known as a random field \cite{KFBook}, defines a probability distribution for a collection of random variables $x\in \mathcal{X}$ with parameter $\theta$ in the form:
\begin{equation}\label{eq:unsup-RF}
p_{\theta}(x)=\frac{1}{Z(\theta)} \exp\left[  u_{\theta}(x) \right] 
\end{equation}
where $\mathcal{X}$ denotes the space of all possible values of $x$, $Z(\theta)=\int\exp\left[  u_{\theta}(x) \right] dx$ is the normalizing constant, $u_{\theta}(x) : \mathcal{X} \rightarrow \mathbb{R}$ is called the potential function which assigns a scalar value to each configuration of $x$ in $\mathcal{X}$ and can be very flexibly defined (e.g., through DNNs of different architectures).
For different applications, $\mathcal{X}$ could be discrete or continuous, and $x$ could be fix-dimensional or trans-dimensional (i.e., sequences of varying lengths). For example, images are fix-dimensional continuous data (i.e., $\mathcal{X}=\mathbb{R}^D$), and natural languages are sequences taking discrete tokens (i.e., $\mathcal{X}=\bigcup_{l}\mathbb{V}^l$ where $\mathbb{V}$ is the vocabulary of tokens).

Training EBMs is challenging, because the gradient in maximizing the data log-likelihood $\log p_\theta({x})$ for observed ${x}$ involves expectation w.r.t. the model distribution $p_\theta(x)$, as shown below:
\begin{equation} \label{eq:RF-grad}
\begin{aligned}
\nabla_\theta \log{p}_{\theta}({x})&=\nabla_\theta u_{\theta}({x})-\nabla_\theta \log Z(\theta)\\
&=\nabla_\theta u_{\theta}({x})-E_{p_\theta(x')}\left[\nabla_\theta u_{\theta}(x')\right].
\end{aligned}
\end{equation}

Considerable progress has been made recently to successfully train large-scale EBMs parameterized by DNNs \cite{nrf,IGG-EBM,jrf,slt} for different types of data from various domains, which lays the foundation to use EBMs, as a unified framework, to achieve domain-agnostic SSL.
\begin{itemize}
	\item For training EBMs for continuous data such as images, the inclusive approach, as detailed in \cite{nrf}, has been shown to yield superior results in unsupervised and semi-supervised training, by introducing inclusive-divergence minimized auxiliary generators and utilizing stochastic gradient sampling (such as SGLD) to approximate the model expectation in Eq. (\ref{eq:RF-grad}).
	\item For training EBMs for discrete sequence data such as natural languages, the DNCE approach, as detailed in \cite{jrf,slt}, avoids the model expectation in Eq. (\ref{eq:RF-grad}) and has achieved superior results in unsupervised and semi-supervised training, with the use of dynamic noise distribution to improve training efficiency of NCE (noise-contrastive estimation) \cite{nce}.
\end{itemize}

\subsection{Pre-training via EBMs for SSL}
\label{sec:pre-training}

Pre-training via EBMs for SSL consists of two stages. The first stage is pre-training an EBM on unlabeled data\footnote{This is also known as unsupervised pre-training, which is different from supervised pre-training (also known as transfer learning using a pre-trained classifier).
It is shown in \cite{oliver2018realistic} that the success of supervised transfer learning heavily
depends on how closely related the two datasets are.
}. This is followed by a fine-tuning stage, where we can easily use the pretrained EBM to initialize a discriminative model and further train over labeled data.

Consider \textbf{pre-training of an EBM for semi-supervised image classification}, which essentially involves estimating $p_\theta(x)$ as defined in Eq.(\ref{eq:unsup-RF}) from unlabeled images.
For the potential function $u_\theta(x)$, we can use a multi-layer feed-forward neural network $\Phi_\theta(x):\mathbb{R}^{D} \rightarrow \mathbb{R}$, which, in the final layer, calculates a scalar via a linear layer, $u_\theta(x) = w^T h$. Here $h \in \mathbb{R}^H$ denotes the activation from the last hidden layer and $w \in \mathbb{R}^H$ the weight vector in the final linear layer. For simplicity, we omit the bias in describing linear layers throughout the paper.

In fine-tuning, we throw $w$ and fed $h$ into an added linear output layer, followed by $softmax(W  h)$, to predict $y$, where $W \in \mathbb{R}^{K \times H}$ denotes the new trainable weight parameters and $y \in \left\lbrace 1,\cdots,K \right\rbrace$ the class label.

The above procedure can be similarly applied to \textbf{pre-training of an EBM for semi-supervised natural language labeling} (e.g., POS tagging). In pre-training, basically we estimate an EBM-based language model $p_\theta(x)$ from unlabeled text corpus.
Neural networks with different architectures can be used to implement the potential function $\Phi_\theta(x):\mathbb{V}^{l} \rightarrow \mathbb{R}$ given length $l$.
With abuse of notation, here $x=(x_1, \ldots, x_l)$ denotes a token sequence of length $l$, and $x_i \in \mathbb{V}, i=1,\cdots,l$.  
We use the bidirectional LSTM based potential function in \cite{slt} as follows:
\begin{equation}\label{eq:u-pretraining-nlp}
u_\theta(x) = \sum_{i=1}^{l-1} h_{f,i}^T e_{i+1} + \sum_{i=2}^{l} h_{b,i}^T e_{i-1}
\end{equation}
where $e_i, h_{f,i}$ and $h_{b,i}$ are of the same dimensions, denoting the output embedding vector, the last hidden vectors of the forward and backward LSTMs respectively at position $i$.

In fine-tuning, we add a CRF, as the discriminative model, on top of the extracted representations $\left\lbrace (h_{f,i}, h_{b,i}), i=1,\cdots,l \right\rbrace $ to do sequence labeling, i.e., to predict a sequence of labels $y=(y_1, \ldots, y_l)$ with one label for one token at each position, where $y_i \in \left\lbrace 1,\cdots,K \right\rbrace$ denotes the label at position $i$.
Specifically, we concatenate $h_{f,i}$ and $h_{b,i}$ and add a linear output layer to define the node potential and add a matrix $A \in \mathbb{R}^{K \times K}$ to define the edge potential, as in recent neural CRFs \cite{lample2016neural,LSTM-CNNs-CRF}. The parameters to be fine-tuned are the weights in the linear output layer and $A$. 

\subsection{Joint-training via EBMs for SSL}
\label{sec:joint-training}

The above pre-training via EBMs for SSL considers the modeling of only observations $x$ without labels $y$.
The joint-training refers to the joint modeling of $x$ and $y$:
\begin{equation}\label{eq:joint-RF}
p_{\theta}(x,y)=\frac{1}{Z(\theta)} \exp\left[  u_{\theta}(x,y) \right] 
\end{equation}
Then,  it can be easily seen that the conditional density $p_{\theta}(y|x)$ implied by the joint density Eq.(\ref{eq:joint-RF}) is:
\begin{equation}\label{eq:rf-classifier}
p_{\theta}(y|x) = \frac{p_{\theta}(x,y)}{p_{\theta}(x)}
= \frac{\exp(  u_{\theta}(x,y) )}{\sum_{y'} \exp(  u_{\theta}(x,y') )}
\end{equation}
The implied marginal density is
$p_\theta(x) = \frac{1}{Z(\theta)} \exp( u_{\theta}(x) )$
where, with abuse of notation, $u_\theta(x) \triangleq log \sum_y \exp\left[  u_{\theta}(x,y) \right]$.
The key for EBM based joint-training for SSL is to choose suitable $u_\theta(x,y)$ such that both $p_\theta(y|x)$ and $p_\theta(x)$ can be tractably optimized.

In \textbf{joint-training of an EBM for semi-supervised image classification}, we consider a neural network $\Psi_\theta(x):\mathbb{R}^{D} \rightarrow \mathbb{R}^K$, which accepts the image $x$ and outputs an vector whose size being equal to the number of class labels, $K$. 
Then we define $u_\theta(x,y) = \Psi_\theta(x)[y]$, where $[y]$ denotes the $y$-th element of a vector.
With the above potential definition, it can be easily seen that the implied conditional density $p_{\theta}(y|x)$ is exactly a standard $K$-class $softmax$ based classifier, using $K$ logits calculated by the neural network $\Psi_\theta(x)$ from the input $x$. And we do not need to calculate $Z(\theta)$ for classification. Therefore, we can conduct SSL over a mix of labeled and unlabeled data by maximizing the (weighted) sum of $\log p_\theta(y|x)$ and $\log p_\theta(x)$, where both optimizations are tractable as detailed in \cite{nrf}.

The above procedure can be similarly applied to \textbf{joint-training of an EBM for semi-supervised natural language labeling} with $x=(x_1, \ldots, x_l)$ and $y=(y_1, \ldots, y_l)$, $x_i \in \mathbb{V}, y_i \in \left\lbrace 1,\cdots,K \right\rbrace, i=1,\cdots,l$. 
We consider a neural network $\Psi_\theta(x): \mathbb{V}^{l} \rightarrow \mathbb{R}^{l \times K}$ and define  
\begin{equation}\label{eq:u-joint-nlp}
u_\theta(x,y) = \sum_{i=1}^l \Psi_\theta(x)[i,y_i] + \sum_{i=1}^l A[y_{i-1},y_i]
\end{equation}
where $[\cdot,\cdot]$ denotes the element of a matrix and  $A \in \mathbb{R}^{K \times K}$ models the edge potential for adjacent labels. With the above potential definition, it can be easily seen that the conditional density $p_{\theta}(y|x)$ implied by the joint density Eq.(\ref{eq:joint-RF}) is exactly a CRF with node potentials $\Psi_\theta(x)[i,y_i]$ and edge potentials $A[y_{i-1},y_i]$, and the implied marginal density $p_{\theta}(x)$ is exactly a trans-dimensional random field (TRF) language model \cite{pami,asru,Bin2018}. Training of both models are tractable as detailed in \cite{jrf,slt}.

\begin{table}
	\centering
	\caption{Applications of EBMs across different domains: comparison and connection (See text for details).}
	\label{tab:application}
	\resizebox{0.5\textwidth}{!}{
		\begin{tabular}{ccc}
			\toprule
			& Image classification         & Natural language labeling     \\
			\midrule
			\multirow{2}{*}{Observation}  & $x\in\mathbb{R}^D$           & $x\in\bigcup_{l}\mathbb{V}^l$ \\
			& continuous, fixed-dimensional & discrete, sequence           \\
			\midrule
			Label                         & $y\in\{1,2,\cdots,K\}$       & $y\in \bigcup_{l}\{1,2,\cdots,K\}^l$      \\
			\midrule
			Pre-training & $u_\theta(x) = w^T h$ & $u_\theta (x)$ in Eq.(\ref{eq:u-pretraining-nlp})                \\
			\midrule
			Joint-training                & $u_\theta(x,y) = \Psi_\theta(x)[y]$             & $u_\theta(x,y)$ in Eq.(\ref{eq:u-joint-nlp})              \\
			\bottomrule
	\end{tabular}}
	\vspace{-2mm}
\end{table}

\section{Experiments}
\label{sec:exp}

SSL experiments are conducted on standard benchmark datasets in different domains, including the CIFAR-10 and SVHN datasets \cite{nrf} for image classification and the POS, chunking and NER datasets \cite{hu2019neural,jrf} for natural language labeling.
We use the standard data split for training and testing.
When we vary the amount of labeled and unlabeled data for training, we select varying proportions (e.g., 10\%, 100\%) of labels from the original full set of labeled data. 
Throughout the paper, the amount of labels is thus described in terms of proportions. ``100\% labeled'' means 50,000 and 73,257 images for CIFAR-10 and SVHN, and 56K, 7.4K, 14K sentences for POS, chunking and NER, respectively.

\subsection{SSL for Image Classification}
\label{sec:SSL-image}

\begin{table}[t]
	\centering
	\vspace{-2mm}
	\caption{SSL for image classification over CIFAR-10 with 4,000 labels.
		The upper/lower blocks show generative/discriminative SSL methods respectively.
		The means and standard deviations are calculated over ten independent runs with randomly sampled labels.
	}
	\label{tab:image-main-result}
	\resizebox{0.48\textwidth}{!}{
		\begin{tabular}{lc}
			\toprule
			Methods                                      & error (\%)                               \\
			\midrule
			CatGAN \cite{springenberg2016unsupervised}   & 19.58$\pm$0.46                           \\
			Ladder network \cite{rasmus2015semi}         & 20.40$\pm$0.47                           \\
			Improved-GAN \cite{imporveGAN}               & 18.63$\pm$2.32                           \\
			BadGAN \cite{badGAN}                         & 14.41$\pm$0.30                           \\
			Sobolev-GAN \cite{sob-gan}                   & 15.77$\pm$0.19                           \\
			\textbf{Supervised baseline}           & 25.72$\pm$0.44                           \\
			\textbf{Pre-training+fine-tuning EBM}        & 21.40$\pm$0.38                                    \\
			\textbf{Joint-training EBM}                  & 15.12$\pm$0.36                           \\
			\midrule
			\multicolumn{2}{c}{Results below this line cannot be directly compared to those above.} \\
			\midrule
			VAT small \cite{miyato2018virtual}           & 14.87                                    \\
			Temporal Ensembling \cite{laine2016temporal} & 12.16$\pm$0.31                           \\
			Mean Teacher \cite{tarvainen2017mean}        & 12.31$\pm$0.28                           \\
			\bottomrule
		\end{tabular}}
\end{table}

First, we experiment with CIFAR-10 and compare different generative SSL methods. As in previous work, we randomly sample 4,000 labeled images for training. The remaining images are treated as unlabeled. We use the network architectures and hyper-parameter settings in \cite{nrf}.
It can be seen from Table \ref{tab:image-main-result} that semi-supervised EBMs, especially the joint-training EBMs, produce strong results on par with state-of-art generative SSL methods\footnote{As discussed in \cite{nrf}, Bad-GANs could hardly be classified as a generative SSL method.}. Furthermore, joint-training EBMs outperform pre-training+fine-tuning EBMs by a large margin in this task.
Note that some discriminative SSL methods, as listed in the lower block in Table \ref{tab:image-main-result}, also produce superior results but heavily utilize domain-specific data augmentations, and thus are not directly compared to generative SSL methods.

\begin{figure}[t]
    \centering
    \includegraphics[width=8.5cm]{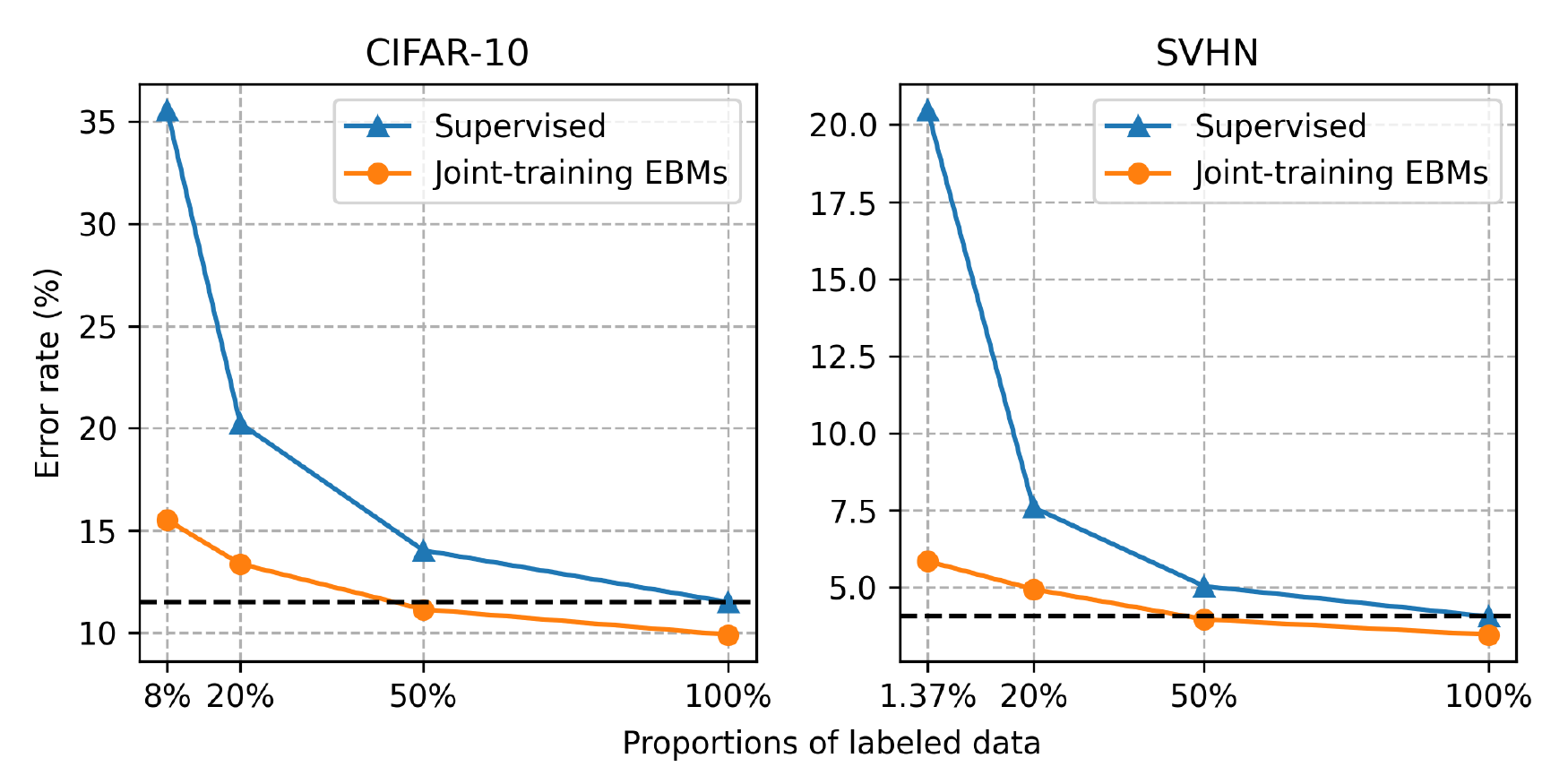}
    \vspace{-3mm}
    \caption{Error rates of supervised baseline and joint-training EBMs as the amount of labels varies on SVHN and CIFAR-10 datasets.
    The dash line is the supervised result trained with 100\% labeled data.}
    \label{fig:sup-semi-compare}
    \vspace{-5mm}
\end{figure}

Second, we experiment with CIFAR-10 and SVHN, and examine the effects of varying amount of labels.
We sample varying proportions of labels as labeled training data and use the remaining as unlabeled training data (i.e., we do not add external unlabeled data).
From the plot of error rates w.r.t. labeling proportions in Fig. \ref{fig:sup-semi-compare} , we can see how many labels can be reduced by using joint-training EBMs. 
The joint-training EBMs obtain 11.14\% on CIFAR-10 and 3.95\% on SVHN using only 50\% labels, which is marginally better than 11.49\% and 4.04\% obtained by the supervised baseline using 100\% labels. This indicates that we can reduce 50\% of labels without losing performance on these two tasks.
Additionally, it is interesting to observe that in the case of using 100\% labels, the joint-training EBMs outperform the supervised baseline with 13.9\% and 14.6\% reductions in error rates. This is because the generative loss $p_\theta(x)$ provides regularization for the pure discriminative loss $p_\theta(y|x)$, as discussed in \cite{ng2002discriminative}.

\subsection{SSL for Natural Language Labeling}
\label{sec:SSL-labeling}

In this experiment, we evaluate different methods for natural language labeling, through three tasks - POS tagging, chunking and NER.
The following benchmark datasets are used - PTB POS tagging, CoNLL-2000 chunking and CoNLL-2003 English NER, as in \cite{LSTM-CNNs-CRF,cvt,hu2019neural,jrf}.
We sample varying proportions of labels as labeled training data and use the Google one-billion-word dataset \cite{google1b} as the large pool of unlabeled sentences.
In \cite{jrf}, joint-training EBM based experiments are conducted, using the labeling proportions of 10\% and 100\% with ``U/L'' (the ratio between the amount of unlabeled and labeled) of 50. In this paper, a larger scale of experiments are conducted, covering the labeling proportions of 2\%, 10\% and 100\% with ``U/L'' of 50, 250 and 500 for three tasks, which consist of 27 settings.
We use the network architectures in \cite{jrf}. After some empirical search, we fix hyper-parameters (tuned separately for different methods), which are used for all the 27 settings. 

From the comparison results in Table \ref{tab:nlp-result} and \ref{tab:nlp-compare}, the main observations are as follows.
1) The joint-training EBMs outperform the supervised baseline in 25 out of the 27 settings. Since we perform one run for each setting, this may indicate 2 outliers.
2) For a fixed labeling size (as given by the labeling proportion), increasing ``U/L'' makes joint-training EBMs performing better, except one outlier.
3) The effect of increasing the labeling size on the improvement of the joint-training EBMs over the supervised baseline with a fixed ``U/L'' is mixed. For POS/chunking/NER, the largest improvements are achieved under 2\%/10\%/100\% labeled, respectively.
It seems that the working point where an SSL method brings the largest improvement over the supervised baseline is task dependent.
If the working point is indicated by the performance of the supervised baseline, then the SSL method brings the largest effect when the performance of the supervised baseline is moderate, neither too low nor already high.
4) Joint-training EBMs outperform pre-training EBMs in 23 out of the 27 settings marginally but nearly consistently.
A possible intuition is that pre-training is not aware of the labels of interest and is thus weakened for representation learning \cite{zoph2020rethinking}.
5) It seems that the degrees of improvements of the joint-training EBMs over the pre-training EBMs are not affected by the labeling size and ``U/L''.

\begin{table}[t]
	\centering
	\caption{
		Natural language labeling results. The evaluation metric is accuracy for POS and $F_1$ for chunking and NER.
		``Labeled''	denotes the amount of labels in terms of the proportions w.r.t. the full set of labels. ``U/L'' denotes the ratio between the amount of unlabeled and labeled data.
		``U/L=0'' denotes the supervised baseline. ``pre.'' and ``joint'' denote the results by pre-training+fine-tuning EBMs and joint-training EBMs, respectively.}
	\label{tab:nlp-result}
	\setlength{\tabcolsep}{1.7mm}{
		\begin{tabular}{c|l|cc|cc|cc}
			\toprule
			\multirow{2}{*}{Labeled} & \multirow{2}{*}{U/L} & \multicolumn{2}{c|}{POS tagging} & \multicolumn{2}{c|}{Chunking} & \multicolumn{2}{c}{NER}                             \\
			                         &                      & pre.                             & joint                        & pre.                      & joint & pre.  & joint \\
			\midrule
			\multirow{4}{*}{$2\%$}   & 0                    & \multicolumn{2}{c|}{95.57}       & \multicolumn{2}{c|}{78.73}    & \multicolumn{2}{c}{78.91}                           \\
			                         & 50                   & 95.72                            & 95.92                         & 81.62                     & 82.24  & 76.74 & 77.61  \\
			                         & 250                  & 95.96                            & 96.13                         & 82.10                     & 82.26  & 78.49 & 78.51  \\
			                         & 500                  & 96.08                            & 96.24                         & 83.10                     & 83.05  & 79.47 & 79.17  \\
			\midrule
			\multirow{4}{*}{$10\%$}  & 0                    & \multicolumn{2}{c|}{96.81}       & \multicolumn{2}{c|}{90.06}    & \multicolumn{2}{c}{86.93}                           \\
			                         & 50                   & 96.87                            & 96.99                         & 91.60                     & 91.85  & 86.37 & 87.05  \\
			                         & 250                  & 96.88                            & 97.00                         & 91.09                     & 91.93  & 86.86 & 86.77  \\
			                         & 500                  & 96.92                            & 97.08                         & 91.93                     & 92.23  & 87.57 & 87.06  \\
			\midrule
			\multirow{4}{*}{$100\%$} & 0                    & \multicolumn{2}{c|}{97.41}       & \multicolumn{2}{c|}{94.77}    & \multicolumn{2}{c}{90.74}                           \\
			                         & 50                   & 97.40                            & 97.49                         & 95.05                     & 95.31  & 91.24 & 91.34  \\
			                         & 250                  & 97.45                            & 97.54                         & 95.12                     & 95.48  & 91.19 & 91.51  \\
			                         & 500                  & 97.46                            & 97.57                         & 95.19                     & 95.50  & 91.30 & 91.52  \\
			\bottomrule
		\end{tabular}}
\end{table}

\begin{table}[t]
	\vspace{-2mm}
	\centering
	\caption{
		Relative improvements by joint-training EBMs compared to the supervised baseline (abbreviated as sup.) and pretraining+fine-tuning EBMs respectively. Refer to Table \ref{tab:nlp-result} for notations.}
	\label{tab:nlp-compare}
	\centering
	\resizebox{0.48\textwidth}{!}{
		\begin{tabular}{c|l|ccc|ccc}
			\toprule
			\multicolumn{2}{c|}{}    & \multicolumn{3}{c|}{joint over sup.} & \multicolumn{3}{c}{joint over pre.}                                           \\
			\midrule
			Labeled                  & U/L                                  & POS                                 & Chunking & NER  & POS & Chunking & NER  \\
			\midrule
			\multirow{3}{*}{$2\%$}   & 50                                   & 7.9                                 & 16.5     & -2.7 & 4.7 & 3.4      & 3.7  \\
			                         & 250                                  & 12.6                                & 16.6     & 1.5  & 4.2 & 0.9      & 0.1  \\
			                         & 500                                  & 15.1                                & 20.3     & 4.5  & 4.1 & -0.3     & -1.5 \\
			\midrule
			\multirow{3}{*}{$10\%$}  & 50                                   & 5.6                                 & 18.0     & 0.9  & 3.8 & 3.0      & 5.0  \\
			                         & 250                                  & 6.0                                 & 18.3     & -1.2 & 3.8 & 9.4      & -0.7 \\
			                         & 500                                  & 8.5                                 & 21.8     & 1.0  & 5.2 & 3.7      & -4.1 \\
			\midrule
			\multirow{3}{*}{$100\%$} & 50                                   & 3.1                                 & 10.3     & 6.5  & 3.5 & 5.3      & 1.1  \\
			                         & 250                                  & 5.0                                 & 13.6     & 8.3  & 3.5 & 7.4      & 3.6  \\
			                         & 500                                  & 6.2                                 & 14.0     & 8.4  & 4.3 & 6.4      & 2.5  \\
			\bottomrule
		\end{tabular}}
	\vspace{-4mm}
\end{table}

\section{Conclusion}
\label{sec:conclusion}

This paper focuses on pushing forward domain-agnostic semi-supervised learning, particularly via energy-based generative models, and makes two contributions. 
First, we explore pre-training via EBMs for SSL and compare it to joint-training.
Second, a suite of experiments are conducted over domains of image classification and natural language labeling to give a realistic whole picture of the performances of EBM based SSL methods.
It is found that joint-training EBMs outperform pre-training EBMs marginally but nearly consistently.
We hope the results presented here make a useful step towards developing domain-agnostic SSL methods.

\vfill\pagebreak


\bibliographystyle{IEEEbib}
\bibliography{refs}

\end{document}